\documentclass{bmvc2k}

\usepackage{xcolor}         
\usepackage{graphicx}
\usepackage{makecell}
\usepackage{adjustbox}
\usepackage{booktabs}
\usepackage{multirow}
\usepackage[table]{xcolor}
\usepackage{bm}
\usepackage{subcaption}
\usepackage{wrapfig} 
\usepackage[most]{tcolorbox}
\usepackage{xcolor}
\usepackage{enumitem}

\usepackage{graphicx}
\usepackage{subcaption}
\usepackage{pifont}
\newcommand{\cmark}{\ding{51}}
\newcommand{\xmark}{\ding{55}}

\newcommand{\methodname}{\textbf{PE-Mamba}}
\usepackage{amsfonts}

\title{PE-Mamba: Bidirectional Selective Layer Aggregation for AI-Generated Image Detection}

\addauthor{Kutub Uddin}{kutub@umich.edu}{1}
\addauthor{Nusrat Tasnim}{tasnim.nishu70@kau.kr}{2}
\addauthor{Khalid Malik}{drmalik@umich.edu}{1}

\addinstitution{
 College of Innovation \& Technology,\\
 University of Michigan-Flint,\\
 Michigan, US
}
\addinstitution{
 School of Electronics and Information Engineering,\\
 Korea Aerospace University,\\
 Goyang, South Korea
}
\runninghead{Uddin et al.}{PE-Mamba for AI-Generated Image Detection}

\def\eg{\emph{e.g}\bmvaOneDot}

\def\etal{\emph{et al}\bmvaOneDot}

\begin{document}

\maketitle
\begin{abstract}
AI-generated image (AIGI) detection has become increasingly challenging due to the rapid advancement of generative models and the diminishing gap
between synthetic and authentic content.
Existing vision transformer-based detectors commonly rely on
weighted-sum strategies to aggregate intermediate
representations across transformer layers, often overlooking the
inherently ordered semantic progression of hierarchical features from
shallow texture cues to deep semantic representations.
In this work, we propose \textbf{PE-Mamba}, a novel framework built upon
a pre-trained PE-Core vision transformer with lightweight LoRA adaptation
that introduces three complementary components for cross-layer
feature aggregation and fusion.
First, a bidirectional selective aggregator (BSA) processes
layer-wise classification tokens through forward and backward selective scans,
where the forward scan progressively accumulates shallow-to-deep forensic
evidence, and the backward scan performs deep-to-shallow contextual
refinement to reinterpret low-level cues in light of high-level semantic
context.
Second, a softmax-weighted aggregator (SWA) computes a learned
global summary of all layer tokens as a complementary
aggregation path.
Third, a sigmoid-gated blend (SGA) adaptively fuses the BSA and SWA
outputs via a learnable scalar gate, allowing the model to dynamically
balance directional sequential evidence and global layer-wise aggregation.
Extensive experiments on UniversalFakeDetect (96.6\% mACC, 99.5\% mAP) and AIGCDetect (95.3\% mACC, 98.1\% mAP) demonstrate that \methodname{} outperforms 18 detectors with superior generalization across diverse generative models, while training only 1.3\% of total parameters (0.13\% for LoRA alone).
\end{abstract}

\section{Introduction} \vspace{-8pt}
\label{sec:intro}
The rapid advancement of generative AI has fundamentally altered the
landscape of digital media and enables the synthesis of photorealistic
images that are increasingly indistinguishable from authentic content.
Generative adversarial networks (GANs)~\cite{karras2019style, uddin2023enhanced, uddin2019anti}, diffusion models~\cite{rombach2022high}, and their derivatives have
demonstrated the capacity to produce high-fidelity images across domains, from natural scenes and human faces to complex
compositions generated from textual prompts~\cite{ramesh2021zero,
nichol2021glide}.
While these technologies enable powerful creative and scientific
applications, they simultaneously pose a profound threat to information
integrity.
Deepfakes have already been implicated in financial fraud exceeding
\$600{,}000 per incident on average~\cite{globenewswire2024, uddin2025adversarial, uddin2022double, farooq2025transferable}, with
projected losses forecast to reach \$40 billion by
2027~\cite{eftsureForecast, uddin2025shield, farooq2026uncertainty}.
These figures underscore the urgent necessity for generalizable
AIGI detectors.\\
Early forensic approaches~\cite{uddin2023robust, uddin2024counter, tasnim2026grex, uddin2025advbench, uddin2025guard, uddin2026face2parts, tasnim2026grex} to detecting synthetic images largely relied on
domain-specific artifacts.
Frequency-domain methods exploit spectral imprints left by upsampling
operations in generative models~\cite{durall2020watch, frank2020leveraging,
tan2024frequency, uddin2023enhanced}, while co-occurrence and gradient-based techniques
capture low-level texture irregularities that distinguish synthetic
from natural images~\cite{nataraj2019detecting, barni2020cnn,
tan2023learning}.
Although effective within their training distribution, such methods
generalize poorly to unseen generative architectures, as each new generative model introduces distinct artifact signatures.
More recent work has shifted toward leveraging large pretrained visual
representations, such as CLIP~\cite{radford2021clip} and
DINO~\cite{caron2021dino}, as universal feature extractors, exploiting
their broad semantic understanding to improve cross-model
generalization~\cite{ojha2023towards, cozzolino2024raising, tan2025c2p}.\\
A particularly promising direction concerns how intermediate representations within vision transformers (ViTs) carry complementary forensic cues across layers.
Shallow transformer layers retain texture and frequency information, while
deeper layers encode semantic and structural
abstractions~\cite{koutlis2024leveraging, liu2024forgery, tasnim2026diversity}.
Prior work aggregating these multi-scale representations via  weighted-average aggregation~\cite{koutlis2024leveraging, uddin2026transformations} demonstrated that exploiting the full hierarchy of transformer features leads to improved detection
performance.
However, existing aggregation strategies typically treat intermediate
features as an unordered set, discarding the meaningful sequential
ordering inherent in the layer hierarchy.\\
In this paper, we introduce \methodname{}, a novel AIGI detection framework that addresses this limitation by modeling cross-layer
dependencies as an ordered sequence using a bidirectional selective state
space models (SSMs).
The proposed approach is motivated by the key observation that layer ordering in a vision transformer is semantically meaningful. The progression from low-level to high-level features constitutes a natural
sequence that directional models are specifically designed to exploit.
Selective SSMs, as introduced in the Mamba
architecture~\cite{gu2023mamba}, generalize classical state space models
with content-adaptive gating.
This selectivity is qualitatively distinct from static weighted-sum aggregation, and is particularly well suited to cross-layer aggregation where layer ordering carries meaningful forensic information for AIGI detection.\\
\methodname{} builds upon PE-Core~\cite{bolya2025perception}, a large-scale vision transformer, and applies lightweight LoRA~\cite {hu2022lora} to the query-key-value (QKV) projections to preserve general visual representations while enabling task-specific tuning.
Forward hooks registered on all second normalization layers extract intermediate classification (CLS) tokens, which are projected to a common dimensionality by a shared projection module and then processed by a bidirectional selective SSM aggregator (BSA), where the forward scan accumulates shallow-to-deep forensic evidence, and the backward scan performs deep-to-shallow contextual reanalysis.
The mean-pooled outputs of both scans are concatenated and adaptively fused with a softmax-weighted global aggregation branch (SWA) via a learned sigmoid gate (SGA), optimized with binary cross-entropy loss.
The main contributions of this work are as follows:\vspace{-6pt}
\begin{itemize}[leftmargin=*, itemsep=2pt]
\item We propose \methodname{}, a novel AIGI detection framework that replaces conventional cross-layer aggregation (learnable weighted average) with a bidirectional selective SSM aggregator (BSA), explicitly modeling the ordered semantic progression of layer-wise CLS token representations in the pretrained vision transformer backbone.
\vspace{-6pt}
\item We introduce a softmax-weighted aggregator (SWA) as a complementary global aggregation branch and a sigmoid gated blend (SGA) that adaptively fuses the directional SSM output with the global summary, enabling more effective fusion of complementary forensic cues across the layer hierarchy.\vspace{-6pt}
\item We demonstrate through extensive experiments that \methodname{} achieves superior detection and cross-model generalization on AIGI benchmarks, while maintaining parameter efficiency through LoRA fine-tuning.
\end{itemize}

\section{Related Work}\vspace{-9pt}
\label{sec:related}
This section reviews prior work across four areas relevant to \methodname{}: 
AIGI detection, SSMs, multi-scale 
feature aggregation, and parameter-efficient fine-tuning. 
\vspace{-15pt}
\subsection{AI-Generated Image Detection}\vspace{-8pt}
Early AIGI detection methods~\cite{tasnim2026comprehensive} focused on identifying statistical artifacts specific to particular generative pipelines. Wang~\etal~\cite{wang2020cnn} demonstrated that CNN-generated images are
detectable due to shared upsampling artifacts, and proposed training on a
single ProGAN~\cite{karras2017progressive} with aggressive data augmentation to improve generalization.
Subsequent work extended this insight to the frequency domain, where GAN
generators leave characteristic spectral footprints arising from periodic
upsampling operations~\cite{durall2020watch, frank2020leveraging,
tan2024frequency}.
Gradient-based representations~\cite{tan2023learning} and NPR~\cite{tan2024rethinking},
which rethinks up-sampling artifacts in CNN-based generative networks,
further characterize low-level texture anomalies that distinguish synthetic
content.
Co-occurrence features~\cite{nataraj2019detecting, barni2020cnn} exploit
pixel-level statistical dependencies as complementary forensic cues.
While effective within their training distribution, these methods exhibit
limited transferability to unseen generative architectures, particularly
as the diversity of generative models has expanded with the proliferation
of diffusion-based methods.
Corvi~\etal~\cite{corvi2023detection} specifically investigated detection
of diffusion-generated images, revealing distinct challenges posed by
score-based synthesis compared to GAN-based generation.\\
The introduction of large pretrained vision-language models as feature extractors substantially improved cross-model generalization.
UniD~\cite{ojha2023towards} demonstrated that CLIP features extracted without fine-tuning provide strong discriminative
signals across a wide range of GAN architectures.
CLIP-based detection was further advanced by
Cozzolino~\etal~\cite{cozzolino2024raising} and the C2PC
framework~\cite{tan2025c2p}, which injects category-common prompts into the CLIP embedding space to improve sensitivity to generation-specific
artifacts.
The DIRE method~\cite{wang2023dire} reconstructs images through diffusion
model inversion to expose diffusion-generated content.
AIDE~\cite{yan2024sanity} proposes a sanity-check evaluation protocol
and reveals that many existing methods overestimate their generalization
capability across diverse generators.
Rajan~\etal~\cite{rajan2024aligned, rajan2025stay} further examine the role of training data alignment and real image feature suppression in
improving detection generalization.\\
Intermediate transformer representations have proven particularly
discriminative for forensic detection.
Koutlis~\etal~\cite{koutlis2024leveraging} demonstrated that
weighted-average aggregation of multi-layer ViT features substantially
outperforms single-layer approaches, establishing hierarchical feature
aggregation as a key design principle that directly motivates our work.
The FatF framework~\cite{liu2024forgery} employs adaptive transformer
blocks that attend to forgery-specific frequency cues extracted from
multiple processing stages.
More recent methods~\cite{yan2024effort, chen2025forgelensdataefficientforgeryfocus, li2025improving, shi2025mirage, li2025towards} push generalization further. EFFT~\cite{yan2024effort} applies orthogonal feature modeling to reduce inter-class interference and showed promising generalization across unseen generative models. ForL~\cite{chen2025forgelensdataefficientforgeryfocus} introduces data-efficient forgery focus based on a weight shared guidance module for improved cross-model transferability. SAFE~\cite{li2025improving} improves generalization by learning transformation-invariant features that remain discriminative under common post-processing operations. MiraGe~\cite{shi2025mirage} constructs multimodal discriminative representations by jointly modeling visual and semantic cues to improve robustness across unseen generators. IAPL~\cite{li2025towards} introduces image-adaptive prompt learning that conditions detection on instance-specific visual characteristics, enabling dynamic adjustment of decision boundaries at inference time.
\vspace{-15pt}
\subsection{Selective State Space Models}\vspace{-8pt}
SSMs have a long history in signal processing and
control theory. For instance, S4~\cite{gu2021efficiently} demonstrated that frequency-domain
parameterization enables SSMs to match attention-based models on
long-sequence tasks with linear complexity.
Mamba~\cite{gu2023mamba} introduced selective state space modeling, where the step size, input, and output matrices are all functions of the input token, enabling content-adaptive gating with linear complexity.
Vision Mamba~\cite{zhu2024vision} and VMamba~\cite{liu2024vmamba}
extended bidirectional SSM processing to vision tasks by scanning image patches in multiple directions to capture spatial dependencies
efficiently.
In this work, we adapt the bidirectional selective SSM paradigm to cross-layer
aggregation of intermediate CLS tokens, applying it to the
depth dimension of the network rather than the spatial token
sequences.
This is well motivated because layer ordering carries structured
semantic meaning, shallow layers encode texture and frequency, deep
layers encode semantics, which SSMs can exploit through their ordered
processing inductive bias.
\vspace{-15pt}
\subsection{Multi-Scale Feature Aggregation}\vspace{-8pt}
Aggregating features across multiple scales or hierarchical levels is
well-established in computer vision.
Feature pyramid networks~\cite{he2016deep} demonstrated the utility of
multi-scale representations for detection and segmentation.
In the context of forgery detection, Koutlis~\etal~\cite{koutlis2024leveraging} systematically studied the use
of intermediate encoder block representations, finding that weighted-average aggregation of multi-layer features substantially outperforms single-layer approaches.
Similarly, the FatF framework~\cite{liu2024forgery} employs adaptive
transformer blocks that attend to forgery-specific frequency features
extracted from multiple processing stages.
Existing multi-scale aggregation strategies for ViT-based detectors
predominantly employ simple learned weighted averaging,
which treats the set of layer representations as an exchangeable
collection without explicit positional bias.
By contrast, the recurrent structure of SSMs provides a hard inductive
bias toward ordered processing, which we argue is particularly appropriate
for the cross-layer aggregation problem, where the directionality of feature transformation from texture to semantics is a meaningful signal that should be explicitly modeled for AIGI detection.
\vspace{-15pt}
\subsection{Parameter-Efficient Fine-Tuning}\vspace{-8pt}
Fine-tuning large pretrained vision models traditionally requires updating all model parameters, which is computationally expensive and risks degrading general representations. LoRA~\cite{hu2022lora} addresses this by injecting trainable low-rank decomposition matrices into specific weight matrices while keeping the original weights frozen, achieving competitive fine-tuning performance with orders of magnitude fewer trainable parameters. In the context of AIGI detection, C2PC~\cite{tan2025c2p} demonstrated that prompt-based fine-tuning of CLIP can substantially improve generalization by injecting category-common knowledge without modifying the backbone weights, highlighting the effectiveness of parameter-efficient adaptation strategies for forensic tasks. Similarly, applying LoRA to the QKV projections of a large pretrained ViT allows the model to adapt its attention patterns to forensic-specific features while preserving the broad visual understanding of the frozen backbone, which is the strategy adopted in \methodname{}.
\vspace{-12pt}
\vspace{-8pt}
\section{Proposed Method}
\begin{figure*}[!t]
\centering
\includegraphics[width=\textwidth]{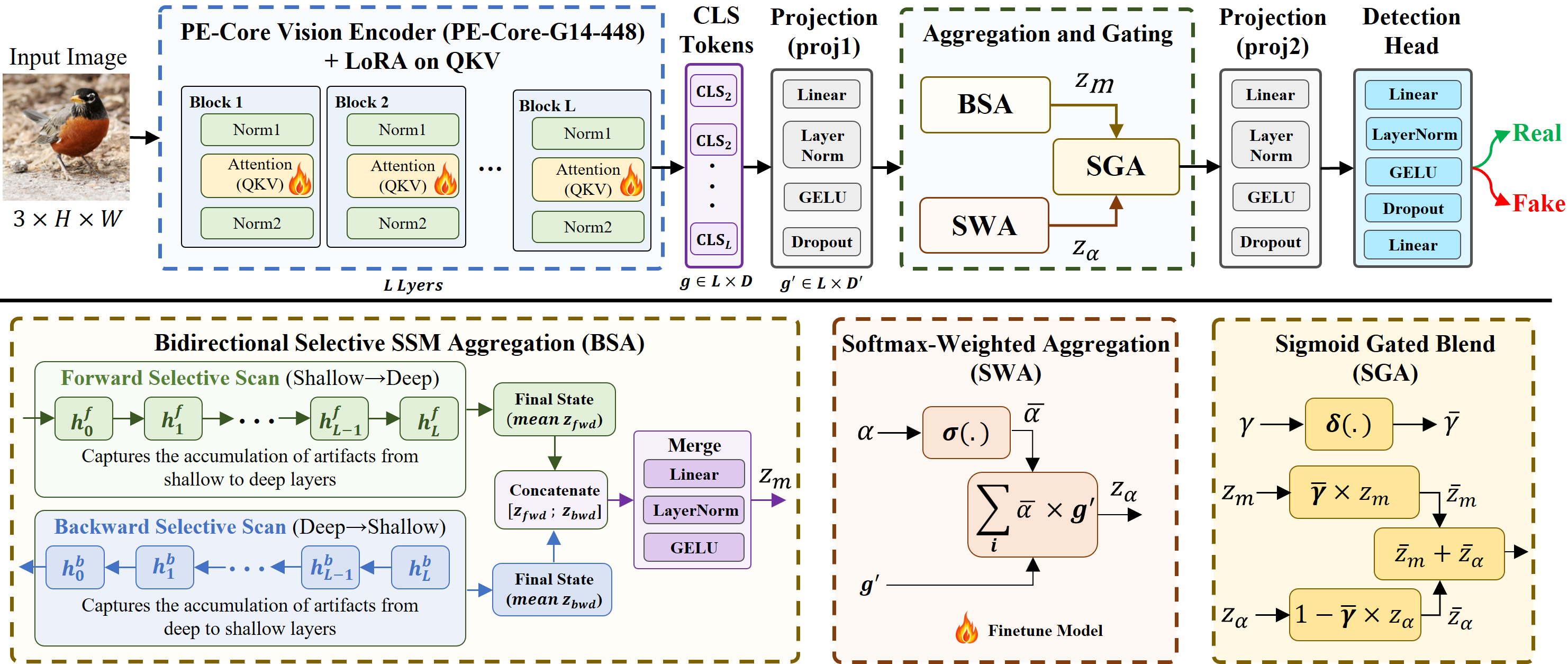}\vspace{8pt}
\caption{Overview of the proposed \methodname{} framework for 
AIGI detection. A frozen PE-Core vision transformer with 
LoRA adaptation on QKV projections extracts layer-wise CLS tokens from 
all second normalization layers, which are projected into a unified embedding 
space by the first projection layer. The BSA processes the projected tokens through forward (shallow to deep) 
and backward (deep to shallow) selective scans to produce $z_m$, while the SWA computes a learned global summary 
$z_{\alpha}$. The SGA adaptively fuses $z_m$ and 
$z_{\alpha}$ via a learnable gate, and the resulting representation is 
refined by the second projection layer and passed to the detection head to detect images as real or AIGI.}
\label{fig:arch}
\vspace{-15pt}
\end{figure*}

\vspace{-12pt}
\label{sec:method}
The overall architecture of \methodname{} is illustrated in
Figure~\ref{fig:arch}.
The framework consists of five principal modules:
(i)~a frozen PE-Core~\cite{bolya2025perception} vision transformer backbone with LoRA adaptation
on the QKV projections;
(ii)~a layer-wise CLS token extraction mechanism via forward hooks;
(iii)~a bidirectional selective SSM aggregator that processes the ordered
sequence of projected CLS tokens;
(iv)~a softmax-weighted aggregator that uses learnable parameters;
and (v)~a sigmoid-gated blend module that blends the SSM output with a learned
alpha-weighted aggregation, followed by a binary detection head.
\vspace{-15pt}
\subsection{Backbone and LoRA Adaptation}\vspace{-8pt}
\methodname{} is built upon PE-Core-G14-448, a gigantic-scale vision
transformer pretrained on large-scale image collections with a patch size
$14 \times 14$ and input resolution $448 \times 448$.
The backbone has embedding dimension $D = 1536$ and consists of $L$
transformer blocks, each containing layer normalization,
multi-head self-attention with QKV projections, a second layer
normalization, and a feed-forward network.
The backbone is initialized from pretrained weights and kept frozen
throughout training.\\
To enable task-specific adaptation without full fine-tuning, we apply
LoRA~\cite{hu2022lora} to the QKV projection matrices of all attention
layers.
For a weight matrix
$\mathbf{W} \in \mathbb{R}^{d_{\text{out}} \times d_{\text{in}}}$,
LoRA introduces trainable matrices
$\mathbf{A} \in \mathbb{R}^{r \times d_{\text{in}}}$ and
$\mathbf{B} \in \mathbb{R}^{d_{\text{out}} \times r}$ with rank
$r \ll \min(d_{\text{out}}, d_{\text{in}})$, replacing the weight update
as $\Delta\mathbf{W} = \mathbf{B}\mathbf{A}$.
We set LoRA rank $r = 8$, scaling factor $\alpha = 8$, and dropout probability $p = 0.1$, applied exclusively to the QKV projection modules.
This restricts the trainable portion of the backbone to fewer than
$0.13\%$ of total parameters with all trainable components (including SSM, projection heads, and gating) comprising 1.3\% in total, preserving the broad visual representations of the pretrained model.
\vspace{-15pt}
\subsection{Layer-wise CLS Token Extraction}\vspace{-8pt}
A key design principle of \methodname{} is the exploitation of intermediate representations across all transformer layers rather than relying solely on the final output token. To achieve this, we register forward hooks on all normalization modules to extract the CLS token after each transformer block and stack them into a tensor $\mathbf{g} \in \mathbb{R}^{L \times D}$, where $L$ is the number of transformer layers and $D = 1536$ is the embedding dimension.\\
Shallow-layer CLS tokens predominantly reflect low-level features, such as local texture, edge statistics, and frequency patterns, cues known to be particularly diagnostic of GAN and diffusion model artifacts~\cite{durall2020watch, tan2023learning}, while deep-layer tokens encode higher-level semantic content capturing structural and compositional properties, making the full sequence $\mathbf{g}$ a rich multi-scale forensic representation whose ordering is semantically meaningful.\\
Before aggregation, each layer's CLS token is projected from $D$ to a common dimension $D'$ through a shared module consisting of a Dropout layer followed by $n$ sequential Linear--LayerNorm--GELU blocks, mapping $\mathbb{R}^{D} \to \mathbb{R}^{D'}$ for the first block and $\mathbb{R}^{D'} \to \mathbb{R}^{D'}$ thereafter, applied identically and independently to each of the $L$ layer tokens to produce $\mathbf{g}' \in \mathbb{R}^{L \times D'}$, with $n = 2$ and $D' = 1024$ in all experiments.
\vspace{-15pt}
\subsection{Bidirectional Selective SSM Aggregator}\vspace{-8pt}
The core contribution of \methodname{} is the bidirectional selective SSM aggregator, which replaces the trainable importance estimation~\cite{koutlis2024leveraging} with a directional, content-adaptive processing module consisting of two independent selective SSM blocks, one for the forward scan and one for the backward scan, followed by a mean-pooling step and a merge module.
\vspace{-15pt}
\subsubsection{Selective SSM Block}\vspace{-8pt}
The selective SSM block~\cite{gu2023mamba} operates on the projected feature sequence $\mathbf{g}' \in \mathbb{R}^{L \times D'}$. An input projection maps $\mathbf{g}'$ to $\mathbb{R}^{L \times 2D_{\text{in}}}$ where $D_{\text{in}} = 2D'$ is the inner dimension, then splits into a main path $\mathbf{x}' \in \mathbb{R}^{L \times D_{\text{in}}}$ and a gating signal $\mathbf{z} \in \mathbb{R}^{L \times D_{\text{in}}}$, followed by a depthwise convolution with kernel size 4 and SiLU activation to provide short-range local context.\\
A linear projection maps $\mathbf{x}'$ to input matrix $\mathbf{B} \in \mathbb{R}^{L \times d_{\text{state}}}$, output matrix $\mathbf{C} \in \mathbb{R}^{L \times d_{\text{state}}}$, and scalar step size $\Delta \in \mathbb{R}^{L \times 1}$, where $d_{\text{state}}=16$ is the SSM state dimension, $\Delta$ is further projected to $D_{\text{in}}$ and activated with Softplus. Since $\mathbf{B}$, $\mathbf{C}$, and $\Delta$ are all functions of the input token, this constitutes the selective mechanism that content-adaptively gates each layer's contribution.\\
The continuous-time state matrix $\mathbf{A} \in \mathbb{R}^{D_{\text{in}} \times d_{\text{state}}}$ is parameterized in log space and held fixed during training, with discrete transition matrix $\overline{\mathbf{A}}_i$ and input matrix $\overline{\mathbf{B}}_i$ computed via zero-order hold, defined as follows:\vspace{-3pt}
\begin{equation}
  \overline{\mathbf{A}}_i = \exp(\Delta_i \odot \mathbf{A}), \quad
  \overline{\mathbf{B}}_i = \Delta_i \odot \mathbf{B}_i
  \label{eq:zoh}
  \vspace{-2pt}
\end{equation}
and the hidden state $\mathbf{h}_i \in \mathbb{R}^{D_{\text{in}} \times d_{\text{state}}}$ updated sequentially over $i = 1, \ldots, L$ as:\vspace{-2pt}
\begin{equation}
  \mathbf{h}_i = \overline{\mathbf{A}}_i \odot \mathbf{h}_{i-1}
               + \overline{\mathbf{B}}_i \odot \mathbf{x}'_i, \quad
  \mathbf{y}_i = \mathbf{h}_i \cdot \mathbf{C}_i^{\top}
  \label{eq:scan}
  \vspace{-2pt}
\end{equation}
where $\odot$ denotes element-wise multiplication and $\mathbf{y}_i \in \mathbb{R}^{D_{\text{in}}}$ is the output at position $i$, giving the scan a hard causal inductive bias unlike learnable weighted average.\\
The scan output $\mathbf{y} \in \mathbb{R}^{L \times D_{\text{in}}}$ is refined via a learned residual skip $\mathbf{y} \leftarrow \mathbf{y} + \mathbf{D} \odot \mathbf{x}'$, where $\mathbf{D} \in \mathbb{R}^{D_{\text{in}}}$ is a learnable skip scale, modulated by output gate $\mathbf{y} \leftarrow \mathbf{y} \odot \text{SiLU}(\mathbf{z})$, projected back to $D'$ via a linear layer, and added to the original input $\mathbf{g}'$ followed by layer normalization, producing output $\mathbf{h}^{f/b} \in \mathbb{R}^{L \times D'}$ for the forward ($f$) or backward ($b$) scan respectively.
\vspace{-10pt}
\subsubsection{Bidirectional Processing and Pooling}\vspace{-8pt}
The bidirectional selective SSM aggregator (BSA) blocks process the projected sequence $\mathbf{g}' \in \mathbb{R}^{L \times D'}$ independently: forward scan processes $\mathbf{g}'$ in natural layer order (Layer~1 to Layer~$L$), accumulating evidence from shallow texture cues toward deep semantic representations, while backward scan processes the reversed sequence of $\mathbf{g}'$, enabling top-down contextual reanalysis where deep semantic information is propagated back to reinterpret low-level forensic features.\\
Both scans produce hidden state sequences of shape $[L, D']$; rather than using only the terminal state, we apply mean pooling over the sequence dimension, defined as follows: \vspace{-5pt}
\begin{equation}
  \mathbf{z}_{\text{fwd}} = \frac{1}{L}\sum_{i=1}^{L}\mathbf{h}^{f}_{i},
  \qquad
  \mathbf{z}_{\text{bwd}} = \frac{1}{L}\sum_{i=1}^{L}\mathbf{h}^{b}_{i}
  \label{eq:pool}
  \vspace{-4pt}
\end{equation}
where $\mathbf{h}^{f}, \mathbf{h}^{b} \in \mathbb{R}^{L \times D'}$ are the forward and backward hidden state sequences, integrating evidence from all layer positions rather than privileging only the terminal state.\\
The two pooled vectors are concatenated and merged:\vspace{-4pt}
\begin{equation}
  \mathbf{z}_{\text{m}} =
    \text{Merge}([\mathbf{z}_{\text{fwd}};\,\mathbf{z}_{\text{bwd}}])
    \in \mathbb{R}^{D'}
  \label{eq:merge}
  \vspace{-4pt}
\end{equation}
where Merge consists of $\text{Linear}(2D' \to D')$, LayerNorm, and GELU.
\vspace{-12pt}
\subsection{Softmax-Weighted Aggregator}\vspace{-8pt}
To complement the sequential inductive bias of the SSM, a parallel softmax-weighted aggregator (SWA) computes a global summary of all projected layer tokens, defined as follows:\vspace{-10pt}
\begin{equation}
  \mathbf{z}_{\alpha} =
    \sum_{\ell=1}^{L}
    \sigma(\boldsymbol{\alpha})_{\ell} \odot \mathbf{g}'_{\ell}
    \in \mathbb{R}^{D'}
  \label{eq:alpha}
  \vspace{-10pt}
\end{equation}
where $\sigma(\cdot)$ denotes the softmax operation and $\boldsymbol{\alpha} \in \mathbb{R}^{L \times D'}$ is a learnable parameter initialized from $\mathcal{N}(0,1)$, providing a global aggregation.
\vspace{-12pt}
\subsection{Sigmoid Gated Blend}\vspace{-8pt}
The BSA output $\mathbf{z}_{\text{m}}$ and the SWA output $\mathbf{z}_{\alpha}$ are adaptively fused via a learned scalar gate $\gamma \in \mathbb{R}$, initialized to zero so that $\delta(\gamma) = 0.5$ gives equal weight to both branches at the start of training, defined as follows:\vspace{-8pt}
\begin{equation}
  \mathbf{z} =
    \delta(\gamma)\,\mathbf{z}_{\text{m}}
    + \bigl(1 - \delta(\gamma)\bigr)\,\mathbf{z}_{\alpha}
    \in \mathbb{R}^{D'}
  \label{eq:gate}
  \vspace{-6pt}
\end{equation}
allowing the model to adaptively emphasize directional sequential evidence or global layer-wise aggregation depending on the training signal. Here, $\delta(\cdot)$ denotes the sigmoid function.\\
The gated representation $\mathbf{z} \in \mathbb{R}^{D'}$ is passed through a second projection module with the same structure as the first projection, a Dropout layer followed by $n$ sequential Linear--LayerNorm--GELU--Dropout blocks operating entirely in $\mathbb{R}^{D'}$. The detection head then applies three Linear--GELU--Dropout blocks followed by a final $\text{Linear}(D' \to 1)$, producing a single logit $p \in \mathbb{R}$ for binary prediction.
\vspace{-10pt}
\section{Results}\vspace{-10pt}
This section presents the experimental setup, quantitative comparisons with 18 state-of-the-art detectors, robustness evaluation under image perturbations, qualitative Grad-CAM analysis, and extensive ablation studies validating the key design choices.
\vspace{-15pt}
\subsection{Implementation Details}\vspace{-8pt}
All experiments are conducted on a Linux 22.04 system equipped with two 48GB GPUs. Training uses AdamW with learning rate $\eta = 10^{-4}$, weight decay $5 \times 10^{-2}$, $\beta_1 = 0.9$, $\beta_2 = 0.999$, and $\varepsilon = 10^{-8}$. Input images are resized to $448 \times 448$ and normalized to ImageNet mean and standard deviation. All experiments use $D' = 1024$, $n = 2$, $d_{\text{state}} = 16$, $d_{\text{expand}} = 2$, LoRA rank $r = 8$, scaling factor $\alpha = 8$, and dropout probability $p = 0.1$ throughout.
\vspace{-15pt}
\subsection{Datasets and Evaluation Protocol}\vspace{-8pt}
\methodname{} trains solely on ProGAN~\cite{karras2017progressive} and evaluates zero-shot on 19 unseen generators spanning three families: GANs (CycleGAN~\cite{zhu2017unpaired}, BigGAN~\cite{brock2018large}, StyleGAN~\cite{karras2019style}, StyleGAN2~\cite{karras2020analyzing}, GauGAN~\cite{park2019semantic}, StarGAN~\cite{choi2018stargan}), other generative models (Deepfakes~\cite{rossler2019faceforensics}, SITD~\cite{chen2018learning}, SAN~\cite{dai2019second}, CRN~\cite{chen2017photographic}, IMLE~\cite{li2019diverse}), and diffusion models (Guided~\cite{dhariwal2021diffusion}, LDM~\cite{rombach2022high}, GLIDE~\cite{nichol2021glide}, DALL-E~\cite{ramesh2021zero}) on the UniversalFakeDetect~\cite{wang2020cnn} benchmark. \\
AIGCDetect~\cite{yan2024sanity} is a fully out-of-distribution benchmark covering nine modern diffusion-based generators: ADM~\cite{dhariwal2021diffusion}, DALL-E~2~\cite{ramesh2022hierarchical}, GLIDE~\cite{nichol2021glide}, Midjourney~\cite{yan2024sanity}, SD~v1.4/v1.5~\cite{rombach2022high}, SD-XL~\cite{podell2024sdxl}, VQDM~\cite{gu2022vector}, and Wukong~\cite{yan2024sanity}, evaluated using the same ProGAN-trained model without any retraining or adaptation. For both benchmarks, we report per-generator ACC and AP along with mACC and mAP across all generators.

\begin{table*}[!t]
\centering
\resizebox{\textwidth}{!}{
\begin{tabular}{c *{6}{c} c *{2}{c} *{2}{c} c *{3}{c} *{3}{c} c c}
\toprule
\midrule
  & \multicolumn{7}{c}{\textbf{GAN Family}}
  & \multicolumn{5}{c}{\textbf{Other Family}}
  & \multicolumn{8}{c}{\textbf{Diffusion Family}} \\
\cmidrule(lr){2-8}
\cmidrule(lr){9-13}
\cmidrule(lr){14-21}
  \rotatebox{90}{Method}
  & \rotatebox{90}{ProGAN}
  & \rotatebox{90}{CycleGAN}
  & \rotatebox{90}{BigGAN}
  & \rotatebox{90}{StyleGAN}
  & \rotatebox{90}{GauGAN}
  & \rotatebox{90}{StarGAN}
  & \rotatebox{90}{Deepfakes}
  & \rotatebox{90}{SITD}
  & \rotatebox{90}{SAN}
  & \rotatebox{90}{CRN}
  & \rotatebox{90}{IMLE}
  & \rotatebox{90}{Guided}
  & \rotatebox{90}{LDM 200}
  & \rotatebox{90}{\makecell{LDM 200\\ CFG}}
  & \rotatebox{90}{LDM 100}
  & \rotatebox{90}{\makecell{Glide 100\\27}}
  & \rotatebox{90}{\makecell{Glide 50\\27}}
  & \rotatebox{90}{\makecell{Glide 100\\10}}
  & \rotatebox{90}{DALL-E}
  & \rotatebox{90}{mACC} \\
\midrule
UpCnv~\cite{durall2020watch}*              & 53.1 & 69.7 & 67.2 & 60.1 & 53.8 & 92.8 & 53.6 & 85.0 & 50.5 & 52.5 & 51.6 & 57.5 & 49.0 & 51.3 & 49.5 & 54.5 & 58.1 & 59.7 & 55.1 & 59.2 \\
CNND~\cite{wang2020cnn}               & 100.0 & 85.2 & 70.2 & 85.7 & 79.0 & 91.7 & 53.5 & 66.7 & 48.7 & 86.3 & 86.3 & 60.1 & 54.0 & 55.0 & 54.1 & 60.8 & 63.8 & 65.7 & 55.6 & 69.6 \\
LGrad~\cite{tan2023learning}          & 99.8 & 85.4 & 82.9 & 94.8 & 72.5 & 99.6 & 58.0 & 62.5 & 50.0 & 50.7 & 50.8 & 77.5 & 94.2 & 95.8 & 94.8 & 87.4 & 90.7 & 89.6 & 88.4 & 80.3 \\
DMD~\cite{corvi2023detection}*        & 100.0 & 91.0 & 96.7 & 99.0 & 91.0 & 99.0 & 67.7 & 94.1 & 56.6 & 99.3 & 99.3 & 53.8 & 57.5 & 59.4 & 57.6 & 56.6 & 58.8 & 58.8 & 69.7 & 77.2 \\
UniD~\cite{ojha2023towards}           & 100.0 & 98.5 & 94.5 & 82.0 & 99.5 & 97.0 & 66.6 & 63.0 & 57.5 & 59.5 & 72.0 & 70.0 & 94.2 & 73.8 & 94.4 & 79.1 & 79.9 & 78.1 & 86.8 & 81.4 \\
NPR~\cite{tan2024rethinking}          & 99.8 & 95.0 & 87.6 & 96.2 & 86.6 & 99.8 & 76.9 & 66.9 & 98.6 & 50.0 & 50.0 & 84.6 & 97.7 & 98.0 & 98.2 & 96.3 & 97.2 & 97.4 & 87.2 & 87.6 \\
AlgnF~\cite{rajan2024aligned}*        & 73.7 & 56.8 & 66.0 & 74.6 & 63.5 & 75.2 & 48.1 & 77.8 & 63.2 & 55.0 & 48.7 & 60.4 & 89.2 & 89.2 & 89.2 & 66.0 & 67.1 & 67.5 & 79.5 & 69.0 \\
FrqNet~\cite{tan2024frequency}        & 97.9 & 95.8 & 90.5 & 97.6 & 50.2 & 93.4 & 97.4 & 88.9 & 59.0 & 71.9 & 67.4 & 86.7 & 84.6 & 99.6 & 65.6 & 85.7 & 97.4 & 88.2 & 59.1 & 83.0 \\
EFFT~\cite{yan2024effort}             & 100.0 & 99.9 & 99.6 & 95.1 & 99.6 & 100.0 & 87.6 & 92.5 & 81.5 & 98.9 & 98.9 & 69.2 & 99.3 & 96.8 & 99.5 & 97.5 & 97.8 & 97.2 & 98.1 & 95.2 \\
FatF~\cite{liu2024forgery}            & 99.9 & 93.3 & 99.5 & 97.2 & 99.4 & 99.8 & 93.2 & 81.1 & 68.0 & 69.5 & 69.5 & 76.0 & 98.6 & 94.9 & 98.7 & 94.4 & 94.7 & 94.2 & 98.8 & 90.6 \\
RINE~\cite{koutlis2024leveraging}     & 100.0 & 99.3 & 99.6 & 88.9 & 99.8 & 99.5 & 80.6 & 90.6 & 68.3 & 89.2 & 90.6 & 76.1 & 98.3 & 88.2 & 98.6 & 88.9 & 92.6 & 90.7 & 95.0 & 91.3 \\
AIDE~\cite{yan2024sanity}            & 99.9 & 98.5 & 83.9 & 99.7 & 73.2 & 99.9 & 54.1 & 68.1 & 71.1 & 60.9 & 61.0 & 88.5 & 98.2 & 97.5 & 98.4 & 98.2 & 98.4 & 97.9 & 97.5 & 86.6 \\
StayP~\cite{rajan2025stay}*           & 54.0 & 51.9 & 54.1 & 59.5 & 54.6 & 79.4 & 53.0 & 55.0 & 68.5 & 50.3 & 49.9 & 51.5 & 98.6 & 98.6 & 97.6 & 62.6 & 65.7 & 64.0 & 91.5 & 66.3 \\
C2PC~\cite{tan2025c2p}               & 99.9 & 97.3 & 99.1 & 96.4 & 99.2 & 99.6 & 93.8 & 95.6 & 64.4 & 93.3 & 93.3 & 69.1 & 99.3 & 97.3 & 99.3 & 95.3 & 95.3 & 96.1 & 98.9 & 93.8 \\
ForL~\cite{chen2025forgelensdataefficientforgeryfocus} & 100.0 & 99.5 & 99.4 & 96.6 & 95.9 & 99.0 & 81.2 & 82.4 & 78.0 & 98.3 & 98.1 & 67.1 & 99.2 & 96.7 & 99.3 & 84.1 & 86.5 & 84.4 & 98.8 & 91.8 \\
SAFE~\cite{li2025improving}            & 99.9 & 98.9 & 89.7 & 98.0 & 91.5 & 99.9 & 93.1 & 85.6 & 95.9 & 50.1 & 50.1 & 82.4 & 98.8 & 98.7 & 98.8 & 95.8 & 96.6 & 97.3 & 97.5 & 90.5 \\
MiraGe~\cite{shi2025mirage}                      & 100.0 & 94.3 & 96.5 & 96.8 & 93.6 & 96.1 & 88.7 & 75.8 & 71.9 & 92.9 & 92.9 & 82.0 & 98.3 & 94.6 & 98.6 & 97.5 & 97.5 & 98.0 & 98.6 & 92.9 \\
IAPL~\cite{li2025towards}                        & 100.0 & 98.6 & 98.7 & 94.9 & 99.4 & 96.7 & 95.9 & 90.8 & 93.8 & 92.5 & 92.7 & 72.8 & 99.5 & 97.7 & 99.2 & 98.0 & 98.3 & 98.4 & 98.9 & 95.6 \\
\midrule
\textbf{PE-Mamba} & 99.9 & 99.9 & 99.9 & 86.6 & 100.0 & 100.0 & 90.4 & 99.4 & 98.2 & 99.8 & 99.8 & 75.5 & 98.1 & 92.7 & 97.9 & 98.9 & 99.2 & 99.5 & 99.0 & \textbf{96.6} \\
\midrule
\bottomrule
\end{tabular}%
}
\vspace{5pt}
\caption{ACC (\%) of all compared detectors on the UniversalFakeDetect~\cite{wang2020cnn} benchmark. All detectors are trained exclusively on ProGAN~\cite{karras2017progressive} (4 categories) and evaluated zero-shot on the remaining generators. * denotes methods retrained on ProGAN for a fair comparison.}
\label{tab:uni_acc}
\vspace{-8pt}
\end{table*}

\begin{table*}[!t]
\centering
\resizebox{\textwidth}{!}{
\begin{tabular}{c *{6}{c} c *{2}{c} *{2}{c} c *{3}{c} *{3}{c} c c}
\toprule
\midrule
  & \multicolumn{6}{c}{\textbf{GAN Family}}
  & \multicolumn{5}{c}{\textbf{Other Family}}
  & \multicolumn{8}{c}{\textbf{Diffusion Family}} \\
\cmidrule(lr){2-7}
\cmidrule(lr){8-12}
\cmidrule(lr){13-20}
  \rotatebox{90}{Method}
  & \rotatebox{90}{ProGAN}
  & \rotatebox{90}{CycleGAN}
  & \rotatebox{90}{BigGAN}
  & \rotatebox{90}{StyleGAN}
  & \rotatebox{90}{GauGAN}
  & \rotatebox{90}{StarGAN}
  & \rotatebox{90}{Deepfakes}
  & \rotatebox{90}{SITD}
  & \rotatebox{90}{SAN}
  & \rotatebox{90}{CRN}
  & \rotatebox{90}{IMLE}
  & \rotatebox{90}{Guided}
  & \rotatebox{90}{LDM 200}
  & \rotatebox{90}{\makecell{LDM 200\\ CFG}}
  & \rotatebox{90}{LDM 100}
  & \rotatebox{90}{\makecell{Glide 100\\27}}
  & \rotatebox{90}{\makecell{Glide 50\\27}}
  & \rotatebox{90}{\makecell{Glide 100\\10}}
  & \rotatebox{90}{DALL-E}
  & \rotatebox{90}{mAP} \\
\midrule
UpCnv~\cite{durall2020watch}*             & 78.8 & 79.3 & 81.9 & 74.7 & 68.6 & 100.0 & 53.5 & 97.1 & 48.0 & 60.1 & 62.5 & 68.7 & 54.2 & 56.7 & 54.9 & 60.7 & 67.0 & 69.1 & 65.5 & 68.5 \\
CNND~\cite{wang2020cnn}               & 100.0 & 93.5 & 84.5 & 99.5 & 89.5 & 98.2 & 89.0 & 73.8 & 59.5 & 98.2 & 98.4 & 73.7 & 70.6 & 71.0 & 70.5 & 80.7 & 84.9 & 82.1 & 70.6 & 83.6 \\
LGrad~\cite{tan2023learning}          & 100.0 & 94.0 & 90.7 & 99.9 & 79.4 & 99.9 & 67.9 & 59.4 & 51.4 & 63.5 & 69.6 & 87.1 & 99.0 & 99.2 & 99.2 & 93.2 & 95.1 & 94.9 & 97.2 & 86.3 \\
UniD~\cite{ojha2023towards}           & 100.0 & 98.1 & 94.5 & 86.7 & 99.3 & 99.5 & 91.7 & 78.5 & 67.5 & 83.1 & 91.1 & 79.2 & 95.8 & 79.8 & 95.9 & 93.9 & 95.1 & 94.6 & 88.5 & 90.1 \\
NPR~\cite{tan2024rethinking}          & 100.0 & 99.5 & 94.5 & 99.9 & 88.8 & 100.0 & 84.4 & 97.9 & 99.9 & 50.2 & 50.2 & 98.3 & 99.9 & 99.9 & 99.9 & 99.9 & 99.9 & 99.9 & 99.3 & 92.8 \\
AlgnF~\cite{rajan2024aligned}*         & 84.1 & 60.4 & 73.3 & 87.4 & 74.0 & 83.9 & 42.8 & 85.8 & 70.7 & 66.6 & 41.7 & 66.6 & 99.9 & 99.9 & 99.9 & 73.5 & 73.6 & 74.4 & 90.3 & 76.3 \\
DMD~\cite{corvi2023detection}*         & 100.0 & 98.3 & 99.7 & 100.0 & 96.8 & 100.0 & 97.5 & 99.9 & 87.2 & 100.0 & 100.0 & 74.5 & 87.6 & 89.2 & 87.8 & 87.4 & 90.3 & 89.4 & 95.7 & 93.8 \\
FrqNet~\cite{tan2024frequency}        & 99.9 & 99.6 & 96.1 & 99.9 & 99.7 & 98.6 & 99.9 & 94.4 & 74.6 & 80.1 & 75.7 & 96.3 & 96.1 & 100.0 & 62.3 & 99.8 & 99.8 & 96.4 & 77.8 & 91.9 \\
EFFT~\cite{yan2024effort}             & 100.0 & 100.0 & 100.0 & 99.8 & 100.0 & 100.0 & 99.0 & 97.5 & 97.5 & 100.0 & 100.0 & 95.4 & 100.0 & 99.9 & 100.0 & 99.9 & 99.9 & 100.0 & 100.0 & 99.4 \\
FatF~\cite{liu2024forgery}            & 100.0 & 100.0 & 99.9 & 99.8 & 100.0 & 100.0 & 97.9 & 97.9 & 81.2 & 99.8 & 99.9 & 91.9 & 99.8 & 99.1 & 99.9 & 99.1 & 99.4 & 99.2 & 99.8 & 98.1 \\
RINE~\cite{koutlis2024leveraging}     & 100.0 & 100.0 & 99.9 & 99.4 & 100.0 & 100.0 & 97.9 & 97.2 & 94.9 & 97.3 & 99.7 & 96.4 & 99.8 & 98.3 & 99.9 & 98.8 & 99.3 & 98.9 & 99.3 & 98.8 \\
AIDE~\cite{yan2024sanity}            & 100.0 & 99.9 & 94.4 & 100.0 & 97.7 & 100.0 & 76.4 & 83.6 & 88.8 & 90.7 & 99.9 & 97.3 & 99.3 & 99.1 & 99.3 & 99.3 & 99.3 & 99.2 & 99.0 & 96.0 \\
StayP~\cite{rajan2025stay}*            & 89.3 & 57.7 & 64.3 & 76.7 & 69.2 & 88.5 & 58.9 & 79.7 & 93.5 & 53.5 & 36.9 & 77.4 & 100.0 & 100.0 & 100.0 & 81.9 & 84.3 & 84.2 & 98.3 & 78.6 \\
C2PC~\cite{tan2025c2p}               & 100.0 & 100.0 & 99.9 & 99.5 & 100.0 & 100.0 & 98.6 & 98.9 & 84.6 & 99.9 & 99.9 & 94.1 & 99.9 & 99.8 & 99.9 & 99.7 & 99.8 & 99.8 & 99.9 & 98.6 \\
ForL~\cite{chen2025forgelensdataefficientforgeryfocus} & 100.0 & 100.0 & 99.9 & 99.2 & 99.5 & 100.0 & 95.1 & 94.7 & 98.6 & 99.8 & 100.0 & 93.3 & 99.9 & 99.8 & 99.9 & 99.7 & 99.8 & 99.9 & 99.9 & 98.9 \\
SAFE~\cite{li2025improving}            & 100.0 & 99.8 & 95.9 & 99.9 & 97.2 & 100.0 & 97.5 & 85.1 & 99.3 & 41.3 & 47.2 & 95.8 & 99.9 & 99.9 & 99.9 & 98.9 & 99.2 & 99.4 & 99.7 & 92.4 \\
MiraGe~\cite{shi2025mirage}                      & 100.0 & 100.0 & 99.9 & 99.8 & 99.9 & 99.9 & 96.0 & 93.9 & 84.7 & 99.9 & 100.0 & 96.4 & 99.9 & 99.1 & 99.9 & 99.8 & 99.7 & 99.8 & 99.9 & 98.3 \\
IAPL~\cite{li2025towards}                          & 100.0 & 99.9 & 99.9 & 99.8 & 100.0 & 100.0 & 97.6 & 97.3 & 98.1 & 100.0 & 100.0 & 96.3 & 99.9 & 99.6 & 99.7 & 99.4 & 99.7 & 99.9 & 99.9 & 99.3 \\
\midrule
\textbf{PE-Mamba} & 100.0 & 100.0 & 99.7 & 99.9 & 99.9 & 99.8 & 98.3 & 97.8 & 100.0 & 99.9 & 99.9 & 97.6 & 99.6 & 99.6 & 99.6 & 99.5 & 99.6 & 99.6 & 99.5 & \textbf{99.5} \\
\midrule
\bottomrule
\end{tabular}%
}\vspace{6pt}
\caption{AP (\%) of all compared detectors on the UniversalFakeDetect~\cite{wang2020cnn} benchmark under the same training and evaluation protocol as Table~\ref{tab:uni_acc}.}
\label{tab:uni_ap}
\vspace{-15pt}
\end{table*}

\begin{table*}[!t]
\centering
\begin{adjustbox}{width=1\textwidth}
\label{tab:acc_ap_combined}
\resizebox{\textwidth}{!}{
\begin{tabular}{lcccccccccc}
\toprule
\midrule
Method & ADM & DALL-E2 & GLIDE & Midjourney & SD-XL & SD v1.4 & SD v1.5 & VQDM & Wukong & mACC/mAP \\
\midrule
UpConv~\cite{durall2020watch}*  & 58.6 / 70.0 & 53.3 / 66.7 & 57.1 / 65.3 & 70.7 / 82.9 & 56.5 / 69.1 & 54.3 / 62.5 & 54.5 / 63.2 & 65.0 / 79.1 & 57.0 / 68.2 & 58.6 / 69.7 \\
CNND~\cite{wang2020cnn}    & 60.2 / 76.2 & 50.4 / 53.6 & 56.9 / 71.5 & 50.8 / 58.7 & 54.9 / 69.8 & 51.1 / 59.2 & 51.2 / 60.0 & 55.1 / 67.6 & 51.1 / 57.0 & 53.5 / 63.7 \\
LGrad~\cite{tan2023learning} & 62.5 / 67.5 & 64.2 / 84.3 & 59.4 / 72.2 & 63.0 / 70.2 & 63.8 / 69.5 & 62.7 / 64.5 & 63.5 / 65.3 & 63.2 / 65.8 & 60.1 / 61.6 & 62.5 / 69.0 \\
UniD~\cite{ojha2023towards} & 66.9 / 87.1 & 50.7 / 63.2 & 62.5 / 84.3 & 56.3 / 74.6 & 60.1 / 88.0 & 63.8 / 86.6 & 63.5 / 86.2 & 85.4 / 96.7 & 71.0 / 91.3 & 64.5 / 84.2 \\
NPR~\cite{tan2024rethinking} & 82.6 / 89.9 & 94.6 / 97.7 & 90.3 / 95.5 & 76.8 / 85.9 & 85.1 / 94.0 & 90.4 / 94.7 & 90.5 / 94.4 & 90.0 / 94.3 & 89.3 / 93.3 & 87.7 / 93.3 \\
AlgnF~\cite{rajan2024aligned}* & 61.7 / 67.7 & 67.8 / 79.3 & 60.3 / 64.3 & 88.8 / 97.1 & 91.2 / 99.3 & 92.0 / 99.7 & 91.7 / 99.0 & 70.3 / 79.5 & 91.3 / 99.9 & 79.5 / 87.3 \\
DMD~\cite{corvi2023detection}* & 52.9 / 72.4 & 50.7 / 74.2 & 58.3 / 86.2 & 52.7 / 74.5 & 53.5 / 82.1 & 54.8 / 82.6 & 54.5 / 82.5 & 61.7 / 90.6 & 59.7 / 86.0 & 55.4 / 81.2 \\
FrqNet~\cite{tan2024frequency} & 78.8 / 85.0 & 59.2 / 61.8 & 77.5 / 83.5 & 62.5 / 66.8 & 80.5 / 87.5 & 85.9 / 91.1 & 85.2 / 90.5 & 77.0 / 83.1 & 85.4 / 90.4 & 76.9 / 82.2 \\
EFFT~\cite{yan2024effort} & 93.7 / 93.7 & 90.7 / 94.8 & 92.2 / 92.0 & 82.1 / 96.8 & 85.4 / 97.9 & 95.3 / 95.2 & 96.1 / 93.8 & 88.7 / 94.7 & 93.3 / 97.8 & 90.8 / 95.2 \\
FatF~\cite{liu2024forgery} & 79.5 / 95.0 & 71.1 / 91.6 & 89.0 / 97.7 & 55.5 / 73.8 & 71.0 / 90.5 & 88.1 / 98.4 & 87.9 / 98.1 & 88.0 / 98.5 & 88.1 / 98.6 & 79.8 / 93.6 \\
RINE~\cite{koutlis2024leveraging} & 74.6 / 96.3 & 54.8 / 89.8 & 80.7 / 97.9 & 57.1 / 87.7 & 64.3 / 96.3 & 83.9 / 98.4 & 83.3 / 98.3 & 89.8 / 99.2 & 84.9 / 98.6 & 74.8 / 95.8 \\
AIDE~\cite{yan2024sanity} & 93.8 / 97.1 & 81.9 / 88.4 & 95.2 / 98.6 & 89.3 / 95.0 & 91.2 / 95.4 & 84.7 / 90.2 & 78.3 / 95.2 & 92.5 / 96.6 & 88.9 / 96.3 & 88.3/94.8\\
StayP~\cite{rajan2025stay}*  & 53.3 / 80.7 & 53.3 / 89.9 & 61.6 / 86.7 & 97.8 / 99.7 & 92.6 / 99.1 & 89.8 / 99.8 & 79.6 / 99.9 & 88.1 / 98.8 & 89.6 / 99.4 & 78.4 / 94.9 \\
C2PC~\cite{tan2025c2p} & 65.1 / 92.2 & 69.6 / 97.3 & 89.0 / 99.0 & 54.9 / 80.4 & 63.7 / 91.4 & 84.2 / 98.7 & 84.3 / 98.6 & 79.1 / 97.9 & 81.9 / 98.2 & 74.6 / 94.9 \\
ForL~\cite{chen2025forgelensdataefficientforgeryfocus} & 79.9 / 96.9 & 87.3 / 98.9 & 95.0 / 99.7 & 58.3 / 82.2 & 73.6 / 91.8 & 94.8 / 99.7 & 94.6 / 99.6 & 86.1 / 98.2 & 92.3 / 99.4 & 84.7 / 96.3 \\
SAFE~\cite{li2025improving} & 85.4 / 91.2 & 92.1 / 99.8 & 91.4 / 95.2 & 87.5 / 92.3 & 76.9 / 84.1 & 85.4 / 96.3 & 81.9 / 96.8 & 84.7 / 94.8 & 87.4 / 93.8 & 85.9 / 93.8 \\
MiraGe~\cite{shi2025mirage} & 91.2 / 96.7 & 86.4 / 90.2 & 92.4 / 94.8 & 85.2 / 94.1 & 86.4 / 93.6 & 93.2 / 97.1 & 89.4 / 95.9 & 90.5 / 96.2 & 93.1 / 97.4 & 89.8 / 95.1 \\
IAPL~\cite{li2025towards} & 90.4 / 95.3 & 89.1 / 94.2 & 94.7 / 97.4 & 93.1 / 97.4 & 91.4 / 95.1 & 90.3 / 97.8 & 93.1 / 98.5 & 88.9 / 96.1 & 92.4 / 96.8 & 91.5 / 96.5 \\
\midrule
\textbf{PE-Mamba} & 86.6 / 93.8 & 97.6 / 99.4 & 99.7 / 99.9 & 78.2 / 90.3 & 99.5 / 99.9 & 99.7 / 100.0 & 99.6 / 99.9 & 98.4 / 99.5 & 99.7 / 99.9 & \textbf{95.3 / 98.1} \\
\midrule
\bottomrule
\end{tabular}
}
\end{adjustbox}
\vspace{5pt}
\caption{ACC / AP (\%) of \methodname{} and competing detectors on the AIGCDetect~\cite{yan2024sanity} benchmark. To assess true generalization, all detectors use the same ProGAN~\cite{karras2017progressive}-trained weights from Tables~\ref{tab:uni_acc} and~\ref{tab:uni_ap} without any retraining or adaptation.}
\label{tab:aigcdetect}
\vspace{-15pt}
\end{table*}

\vspace{-12pt}
\subsection{Results on UniversalFakeDetect Benchmark Dataset}\vspace{-8pt}
Tables~\ref{tab:uni_acc} and~\ref{tab:uni_ap} report ACC and AP across all 19 generators under the UniversalFakeDetect benchmark, where detectors are trained solely on ProGAN and evaluated zero-shot on unseen generators. \methodname{} achieves the best overall performance with 96.6\% mACC and 99.5\% mAP, consistently outperforming prior detectors across GAN, diffusion, and other generator families. These results indicate strong cross-generator generalization and reliable score calibration despite substantial shifts in generation mechanisms.\\
Among GAN-based generators, early artifact-oriented methods, including CNND~\cite{wang2020cnn} and LGrad~\cite{tan2023learning}, perform competitively on generators that preserve low-level synthesis artifacts, such as ProGAN, StyleGAN, and StarGAN, but generalize poorly to unseen distributions and non-GAN families. For example, CNND drops to 48.7\% ACC on SAN and 53.5\% on Deepfakes, while LGrad exhibits substantial degradation on SITD (62.5\%) and SAN (50.0\%). This behavior suggests an over-reliance on generator-specific frequency or upsampling artifacts that do not transfer across synthesis paradigms.\\
Recent methods designed for stronger generalization, such as NPR~\cite{tan2024rethinking}, FrqNet~\cite{tan2024frequency}, EFFT~\cite{yan2024effort}, and RINE~\cite{koutlis2024leveraging}, achieve improved robustness but remain inconsistent across families. Although FrqNet attains strong AP values, its ACC collapses on GauGAN (50.2\%) and SAN (59.0\%), revealing unstable decision boundaries despite confident predictions. Likewise, RINE achieves competitive overall performance but exhibits notable weaknesses on SAN (68.3\%) and SITD (90.6\%), suggesting reduced generalization under semantic or structural distribution shifts.\\
In contrast, \methodname{} demonstrates highly stable behavior across all generator families. Within GAN-based generators, it achieves near-perfect detection on ProGAN, GauGAN, and StarGAN ($\geq$99.9\% ACC) while preserving strong performance on more challenging domains, such as Deepfakes (90.4\%). More importantly, unlike many prior methods that deteriorate under diffusion-based synthesis, \methodname{} maintains consistently strong results across diffusion generators, including Glide, LDM variants, and DALL-E, achieving accuracies ranging from 97.9\% to 99.5\%. This stability highlights the ability of the proposed directional selective SSM aggregation to capture transferable structural inconsistencies beyond generator-specific artifacts.\\
Compared with the strongest prior competitors, \methodname{} surpasses RINE by +5.3\% mACC and +0.7\% mAP while also outperforming EFFT (95.2\% mACC) and IAPL (95.6\% mACC). Particularly large gains are observed on SAN (98.2\% vs.\ 68.3\% for RINE) and SITD (99.4\% vs.\ 90.6\% for RINE), suggesting that directional state-space aggregation better models long-range dependencies and spatial relationships than weighted average mechanisms.\\
The AP results in Table~\ref{tab:uni_ap} further confirm the generalization of \methodname{}. The proposed method achieves 99.5\% mAP and maintains $\geq$99.5\% AP on 17 out of 19 generators, exceeding EFFT (99.4\%) and RINE (98.8\%). Unlike methods that exhibit high AP but unstable classification accuracy, \methodname{} simultaneously achieves strong ranking quality and consistent decision boundaries, indicating well-calibrated feature representations across diverse unseen generation processes.
\vspace{-15pt}
\subsection{Results on AIGCDetect Benchmark Dataset}
\vspace{-8pt}
Table~\ref{tab:aigcdetect} reports ACC/AP on the AIGCDetect benchmark, where all detectors are evaluated zero-shot using the same ProGAN-trained weights, providing a strict test of out-of-distribution generalization to modern high-fidelity generators.\\
\methodname{} achieves 95.3\% mACC and 98.1\% mAP, ranking first among all 18 methods, and representing a substantial gain over the closest competitors IAPL~\cite{li2025towards} (91.5\%/96.5\%) and EFFT~\cite{yan2024effort} (90.8\%/95.2\%).\\
Methods that perform well on UniversalFakeDetect suffer significant drops on AIGCDetect: RINE~\cite{koutlis2024leveraging} falls to 74.8\% mACC and FatF~\cite{liu2024forgery} to 79.8\%, degradations of over 10 points, whereas \methodname{} maintains 95.3\% mACC, only 1.3 points below its UniversalFakeDetect score, demonstrating substantially better generalization. \methodname{} achieves near-perfect detection on SD~v1.4 (99.7\%), SD~v1.5 (99.6\%), SD-XL (99.5\%), GLIDE (99.7\%), and Wukong (99.7\%), outperforming all competing detectors.\\
Midjourney (78.2\% ACC / 90.3\% AP) and ADM (86.6\% ACC / 93.8\% AP) remain the most challenging generators, as Midjourney's proprietary aesthetic refinement pipeline suppresses standard synthesis artifacts, making fake images visually closer to real photographs, while ADM's classifier-guided synthesis produces artifact profiles that differ substantially from both the ProGAN-based training distribution and standard latent diffusion generators. Notably, while StayP~\cite{rajan2025stay} achieves the highest Midjourney ACC (97.8\%), it does so by suppressing real image features entirely, a fundamentally different strategy that trades off performance on other generators, yielding only 78.4\% mACC overall. \methodname{} achieves a better balance, with competitive Midjourney AP (90.3\%) and strong performance across all remaining generators.\\
The most meaningful comparison is with RINE~\cite{koutlis2024leveraging}, which uses the same cross-layer ViT feature aggregation paradigm but relies on a weighted sum with trainable importance weights. \methodname{} surpasses RINE by +20.5\% mACC and +2.3\% mAP, directly validating that replacing learnable weighted average with a bidirectional selective SSM provides substantially more generalizable forensic representations for out-of-distribution generators.
\begin{table*}[!t]
\vspace{6pt}
\centering
\resizebox{\textwidth}{!}{
\begin{tabular}{l *{3}{c} *{3}{c} *{3}{c} *{4}{c} c}
\toprule
\midrule
  & \multicolumn{3}{c}{\textbf{JPEG Compression}}
  & \multicolumn{3}{c}{\textbf{Gaussian Blur}}
  & \multicolumn{3}{c}{\textbf{Additive Noise}}
  & \multicolumn{4}{c}{\textbf{Combinations}} \\
\cmidrule(lr){2-4}
\cmidrule(lr){5-7}
\cmidrule(lr){8-10}
\cmidrule(lr){11-14}
  & \rotatebox{90}{$q$=90}
  & \rotatebox{90}{$q$=75}
  & \rotatebox{90}{$q$=65}
  & \rotatebox{90}{$\sigma$=0.5}
  & \rotatebox{90}{$\sigma$=1.0}
  & \rotatebox{90}{$\sigma$=2.0}
  & \rotatebox{90}{$\sigma$=0.01}
  & \rotatebox{90}{$\sigma$=0.03}
  & \rotatebox{90}{$\sigma$=0.05}
  & \rotatebox{90}{\makecell{JPEG-75\\+Blur-1.0}}
  & \rotatebox{90}{\makecell{JPEG-75\\+Noise-0.03}}
  & \rotatebox{90}{\makecell{Blur-1.0\\+Noise-0.03}}
  & \rotatebox{90}{\makecell{JPEG-75+\\Blur-1.0+\\Noise-0.03}}
  & \rotatebox{90}{\textbf{Clean}} \\
\midrule
mACC & 92.5 & 91.2 & 88.3 & 92.4 & 93.6 & 86.3 & 90.2 & 89.3 & 88.0 & 90.8 & 90.6 & 90.0 & 88.4 & \textbf{96.6} \\
mAP  & 96.2 & 96.4 & 94.3 & 97.6 & 97.5 & 96.1 & 95.9 & 94.4 & 93.1 & 97.7 & 95.9 & 95.4 & 94.5 & \textbf{99.5} \\
\midrule
\bottomrule
\end{tabular}
}
\vspace{6pt}
\caption{Robustness of \methodname{} to JPEG compression ($q \in \{90, 75, 65\}$), Gaussian blur ($\sigma \in \{0.5, 1.0, 2.0\}$), additive noise ($\sigma \in \{0.01, 0.03, 0.05\}$), and their combinations on UniversalFakeDetect~\cite{wang2020cnn} benchmark dataset.}
\label{tab:perturbation}
\vspace{-12pt}
\end{table*}
\vspace{-15pt}
\subsection{Robustness to Image Perturbations}
\vspace{-8pt}
To evaluate the robustness of \methodname{} under real-world post-processing conditions, we apply three common perturbation types~\cite{uddin2023deep}, JPEG compression~\cite{uddin2021analysis}, Gaussian blur~\cite{tasnim2025ai}, and additive Gaussian noise~\cite{tasnim2025ai}, at multiple severity levels, as well as their combinations, to the UniversalFakeDetect benchmark, as reported in Table~\ref{tab:perturbation}.\\
\methodname{} degrades gracefully under mild to moderate perturbations, retaining 92.5\% mACC under JPEG $q$=90 and 93.6\% under Blur $\sigma$=1.0, indicating that the bidirectional cross-layer representations are not overly sensitive to minor signal distortions. The most challenging single perturbation is Blur $\sigma$=2.0 (86.3\% mACC), which significantly smooths high-frequency texture artifacts that are particularly diagnostic for GAN-generated content~\cite{durall2020watch}, followed by JPEG $q$=65 (88.3\%) and Noise $\sigma$=0.05 (88.0\%). Notably, AP remains consistently high ($\geq$93.1\%) across all single-perturbation conditions, indicating well-calibrated confidence scores even when accuracy degrades.\\
Under the most challenging triple combination of JPEG-75 + Blur-1.0 + Noise-0.03, \methodname{} retains 88.4\% mACC and 94.5\% mAP, a degradation of only 8.2 points in mACC from the baseline (96.6\%), demonstrating that the model's representations are sufficiently distributed across multiple cues to defend simultaneous multi-domain distortion.\\
These results confirm that \methodname{} maintains strong performance across a broad range of post-processing conditions without any adversarial training, suggesting that the hierarchical cross-layer aggregation strategy captures forensic evidence at multiple abstraction levels, making it inherently more robust than detectors that rely on a single discriminative signal.
\begin{figure*}[!t]
\centering
\includegraphics[width=\textwidth]{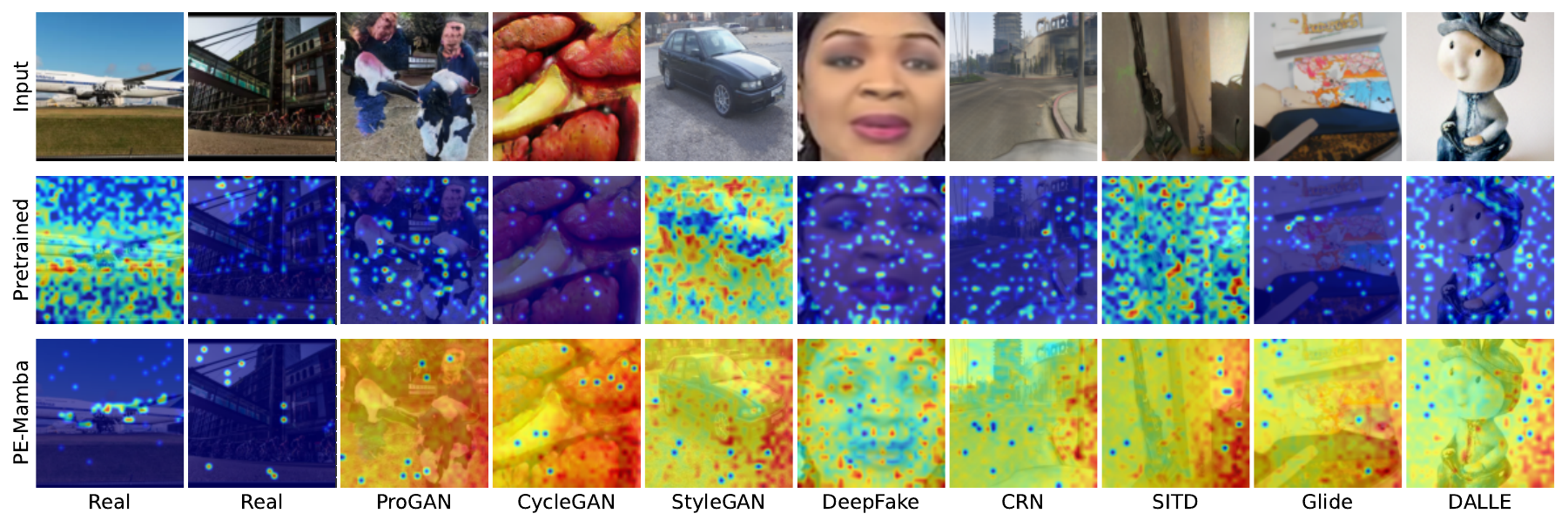}\vspace{6pt}
\caption{Grad-CAM visualizations comparing the pretrained PE-Core and the proposed \methodname{} across real and AIGIs from diverse generators. While the pretrained model produces 
diffuse, semantically driven activations regardless of image 
authenticity, \methodname{} learns to suppress activations on real 
content and concentrate on the forensically discriminative regions in 
fake images, including object boundaries in GAN-generated images, 
facial artifacts in Deepfakes, and global synthesis patterns in 
diffusion-generated content.}
\label{fig:gradcam}
\vspace{-15pt}
\end{figure*}
\vspace{-15pt}
\subsection{Qualitative Analysis via Grad-CAM}
\vspace{-8pt}
To provide interpretable insight into the decision-making of \methodname{}, we visualize class activation maps using Grad-CAM~\cite{selvaraju2017gradcam} for both the pretrained PE-Core and \methodname{} across real and eight generators, as shown in Figure~\ref{fig:gradcam}.\\
The pretrained model produces diffuse, semantically driven activations that are largely consistent regardless of whether the image is real or synthetic. Activations are scattered across salient semantic regions, faces, objects, and textures, reflecting the model's general visual understanding rather than any forensic specialization. Crucially, the pretrained backbone cannot distinguish between real and fake images at the activation level, as evidenced by similarly structured heatmaps across all columns.\\
In contrast, \methodname{} learns distinctly different activation patterns: for real images (columns 1--2), activations are sparse and suppressed, while for fake images (columns 3--10), \methodname{} concentrates on generator-specific forensic regions, localized boundary artifacts in ProGAN; global style-transfer patterns in CycleGAN; structural edge artifacts in StyleGAN; facial skin texture and lip regions in DeepFake; wide-field synthesis patterns in CRN and SITD; and structured contours in diffusion-generated content (Glide, DALL-E).
These visualizations confirm that bidirectional cross-layer aggregation successfully specializes the representation toward artifact-discriminative features without any explicit spatial supervision, demonstrating that ordered shallow-to-deep and deep-to-shallow scanning internalizes forensically meaningful spatial patterns across the full transformer hierarchy.

\begin{table*}[t]
\centering
\setlength{\tabcolsep}{4pt}
\begin{minipage}{0.36\textwidth}
  \centering
  \resizebox{\linewidth}{!}{
  \begin{tabular}{ccc}
    \toprule
    \midrule
    \textbf{SSM Layers} & \textbf{mACC (\%)} & \textbf{mAP (\%)} \\
    \midrule
    \textbf{1}  & \textbf{96.6} & \textbf{99.5} \\
    2           & 95.5    & 99.0   \\
    3           & 96.0    & 99.3    \\
    4           & 95.6    & 98.9    \\
    \midrule
    \bottomrule
  \end{tabular}
  }
  \vspace{6pt}
  \captionof{table}{Number of SSM layers.}
  \label{tab:ssm_layer}
  \vspace{-12pt}
\end{minipage}
\hfill
\begin{minipage}{0.29\textwidth}
  \centering
  \resizebox{\linewidth}{!}{
  \begin{tabular}{ccc}
    \toprule
    \midrule
    \textbf{$d_{\text{state}}$} & \textbf{mACC (\%)} & \textbf{mAP (\%)} \\
    \midrule
    8            & 95.8    & 99.0    \\
    \textbf{16}  & \textbf{96.6} & \textbf{99.5} \\
    32           & 96.3    & 99.3    \\
    64           & 95.9    & 99.1    \\
    \midrule
    \bottomrule
  \end{tabular}
  }
  \vspace{6pt}
  \captionof{table}{State dimension.}
  \label{tab:ssm_state}
  \vspace{-12pt}
\end{minipage}
\hfill
\begin{minipage}{0.31\textwidth}
  \centering
  \resizebox{\linewidth}{!}{
  \begin{tabular}{ccc}
    \toprule
    \midrule
    \textbf{$d_{\text{expand}}$} & \textbf{mACC (\%)} & \textbf{mAP (\%)} \\
    \midrule
    1           & 95.7    & 98.9    \\
    \textbf{2}  & \textbf{96.6} & \textbf{99.5} \\
    3           & 96.4    & 99.3    \\
    4           & 96.1    & 99.1    \\
    \midrule
    \bottomrule
  \end{tabular}
  }
  \vspace{6pt}
  \captionof{table}{SSM expansion.}
  \label{tab:ssm_expand}
  \vspace{-12pt}
\end{minipage}
\end{table*}

\begin{figure}[!t]
\vspace{4pt}
\centering
\begin{minipage}{0.32\linewidth}
    \centering
    \label{fig:abl_lora_r}
    \includegraphics[width=\linewidth]{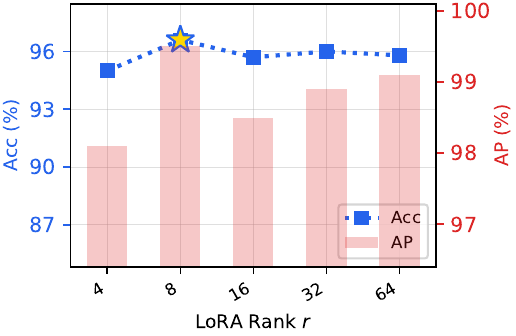}
    \vspace{-10pt}
    (a)
\end{minipage}
\hfill
\begin{minipage}{0.32\linewidth}
    \centering
    \label{fig:abl_lora_a}
    \includegraphics[width=\linewidth]{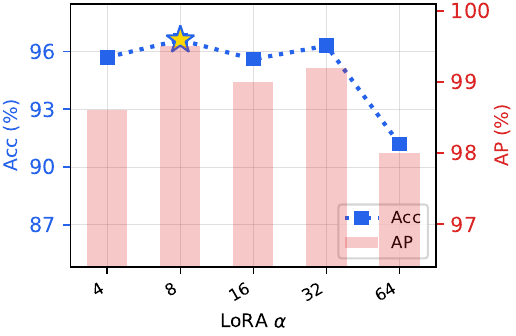}
    \vspace{-10pt}
    (b)
\end{minipage}
\hfill
\begin{minipage}{0.32\linewidth}
    \centering
    \label{fig:abl_lora_d}
    \includegraphics[width=\linewidth]{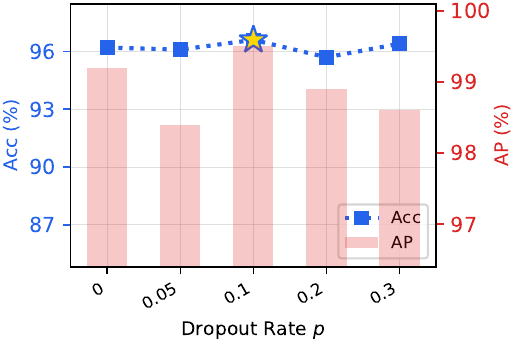} 
    \vspace{-10pt}
    (c)
\end{minipage}
\vspace{16pt}
\caption{Ablation on LoRA hyperparameters: (a) rank $r$, (b) scaling factor $\alpha$, and (c) dropout probability $p$. Gold stars mark the best ACC.}
\label{fig:abl_lora}
\vspace{-12pt}
\end{figure}
\vspace{-12pt}
\subsection{Ablation Study}\vspace{-8pt}
We conduct an extensive ablation study to validate the key design choices of \methodname{}, covering SSM configurations (SSM layers, $d_{\text{state}}$, $d_{\text{expand}}$), LoRA adaptation ($r$, $\alpha$, $p$), aggregation components (BSA, SWA, SGA), backbone selection, projection dimension $D'$, and number of projection layers $n$. All experiments follow the standard ProGAN-trained protocol on UniversalFakeDetect~\cite{wang2020cnn}, with a single variable changed at a time while all other settings are fixed to the defaults ($r{=}8$, $\alpha{=}8$, $p{=}0.1$, $D'{=}1024$, $n{=}2$, $d_{\text{state}}{=}16$, $d_{\text{expand}}{=}2$).
\vspace{-12pt}
\subsubsection{SSM Hyperparameter Analysis}\vspace{-8pt}
We first ablate three SSM-specific hyperparameters, number of SSM layers, state dimension $d_{\text{state}}$, and expansion factor $d_{\text{expand}}$, to justify the configuration used in \methodname{}.\\
Table~\ref{tab:ssm_layer} shows that a single SSM layer achieves the best performance (96.6\% mACC / 99.5\% mAP), with deeper stacking providing no benefit and slightly degrading accuracy. This suggests that the cross-layer forensic aggregation task is well-suited to a single sequential pass, the bidirectional forward and backward scans already provide sufficient representational depth without requiring additional stacked recurrence, and deeper stacking likely introduces over-smoothing across the $L$ layer positions.\\
Table~\ref{tab:ssm_state} evaluates $d_{\text{state}} \in \{8, 16, 32, 64\}$, where $d_{\text{state}} = 16$ achieves the best result (96.6\% mACC / 99.5\% mAP). Smaller values ($d_{\text{state}} = 8$) slightly limit the hidden state capacity (95.8\% mACC), while larger values ($d_{\text{state}} \geq 32$) do not improve performance, indicating that the forensic information compressed from $L$ layers is low-dimensional and well-captured by a state of size 16.\\
Table~\ref{tab:ssm_expand} shows that $d_{\text{expand}} = 2$ achieves the best trade-off (96.6\% mACC / 99.5\% mAP). A smaller expansion ($d_{\text{expand}} = 1$) reduces the inner representation capacity (95.7\% mACC), while larger values ($d_{\text{expand}} \geq 3$) offer marginal gains at increased computational cost. The standard Mamba expansion factor of 2 proves optimal for this task, consistent with its effectiveness in the original sequence modeling literature~\cite{gu2023mamba}.
\vspace{-12pt}
\subsubsection{LoRA Hyperparameter Ablation} \vspace{-8pt}
\label{sec:ablation:lora}
We ablate the three key LoRA hyperparameters, rank $r$, scaling factor $\alpha$, and dropout rate $p$, while keeping all other settings fixed.
Results are summarized in Figures~\ref{fig:abl_lora}.\\
As shown in Figure~\ref{fig:abl_lora}(a), $r{=}8$ achieves the best trade-off (96.6\% mACC, 99.5\% mAP). Lower rank ($r{=}4$) slightly limits expressiveness (95.0\% mACC), while higher ranks ($r{\geq}16$) do not improve and marginally reduce accuracy, likely due to overfitting to the training domain. The narrow performance gap across $r \in \{8, 16, 32, 64\}$ suggests that the task-relevant adaptation signal is low-dimensional and well-captured by a small rank.\\
Figure~\ref{fig:abl_lora}(b) shows that performance peaks at $\alpha{=}8$ with 96.6\% mACC and 99.5\% mAP. Smaller values (\eg $\alpha{=}4$) slightly underperform (95.7\% mACC), while larger values degrade more noticeably ($\alpha{=}64$ drops to 91.2\% mACC and 98.0\% mAP), suggesting that an overly large scaling factor destabilizes fine-tuning by amplifying LoRA updates beyond the pretrained feature distribution.\\
Figure~\ref{fig:abl_lora}(c) shows the model is largely robust to dropout rate, with all settings achieving between 95.7\% and 96.6\% mACC. A moderate dropout of $p{=}0.1$ yields the best result (96.6\% mACC, 99.5\% mAP), providing a consistent regularization benefit, while both zero dropout and higher rates (\eg $p{=}0.3$) perform comparably, indicating that the low-rank constraint provides sufficient implicit regularization.\\
Based on these results, we adopt $r{=}8$, $\alpha{=}8$, and $p{=}0.1$ as the final LoRA configuration.

\begin{table*}[t]
\centering
\setlength{\tabcolsep}{4pt}
\begin{minipage}{0.44\textwidth}
  \centering
  \resizebox{\linewidth}{!}{
  \begin{tabular}{ccccc}
    \toprule
    \midrule
    \textbf{BSA} & \textbf{SWA} & \textbf{SGA} & \textbf{mACC  (\%)} & \textbf{mAP (\%)} \\
    \midrule
    \xmark & \xmark & \xmark & 90.6 & 96.9 \\
    \cmark & \xmark & \xmark & 95.3 & 98.7 \\
    \xmark & \cmark & \xmark & 95.1 & 98.3 \\
    \cmark & \cmark & \xmark & 96.1 & 99.0 \\
    \midrule
    \cmark & \cmark & \cmark & \textbf{96.6} & \textbf{99.5} \\
    \midrule
    \bottomrule
  \end{tabular}
  }
  \vspace{6pt}
  \captionof{table}{Aggregation components.}
  \label{tab:ablation:arch}
  \vspace{-12pt}
\end{minipage}
\hspace{0.01\textwidth}
\begin{minipage}{0.44\textwidth}
  \centering
  \resizebox{\linewidth}{!}{
  \begin{tabular}{lcc}
    \toprule
    \midrule
    \textbf{Backbone} & \textbf{mACC (\%)} & \textbf{mAP (\%)} \\
    \midrule
    DINOv2-L14-448  & 91.7 & 98.3 \\
    CLIP-ViT-L14-224    & 94.8 & 98.5 \\
    CLIP-ViT-L14-336    & 93.6 & 98.4 \\
    PE-Core-L14-336     & 95.7 & 98.9 \\
    \midrule
    \textbf{PE-Core-G14-448}  & \textbf{96.6} & \textbf{99.5} \\
    \midrule
    \bottomrule
  \end{tabular}
  }
  \vspace{6pt}
  \captionof{table}{Different backbone selection.}
  \label{tab:backbone}
  \vspace{-12pt}
\end{minipage}
\end{table*}

\begin{figure}[!t]
\vspace{4pt}
\centering
\begin{minipage}{0.32\linewidth}
    \centering
    \label{fig:abl_lora_r}
    \includegraphics[width=\linewidth]{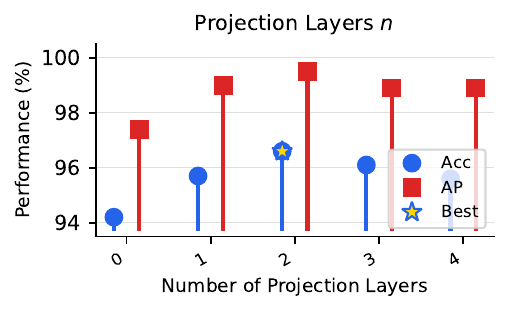}
    \vspace{-10pt}
    (a)
\end{minipage}
\hfill
\begin{minipage}{0.32\linewidth}
    \centering
    \label{fig:abl_lora_a}
    \includegraphics[width=\linewidth]{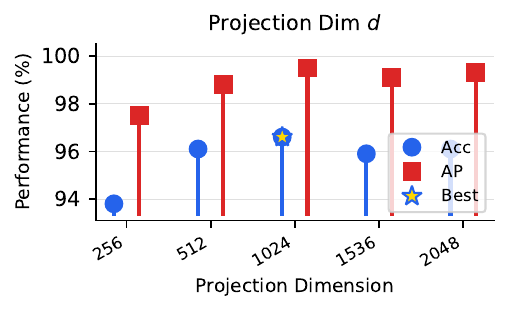}
    \vspace{-10pt}
    (b)
\end{minipage}
\hfill 
\begin{minipage}{0.32\linewidth}
    \centering
    \label{fig:abl_lora_d}
    \includegraphics[width=\linewidth]{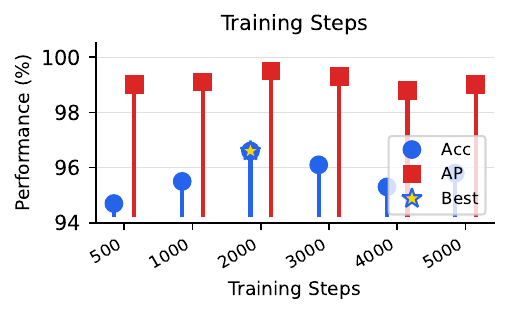} 
    \vspace{-10pt}
    (c)
\end{minipage}
\vspace{16pt}
\caption{
    Ablation on (a) projection layers $n$, (b), projection dimension $d$, and (c) training steps.
    Optimal settings are 2{,}000 steps, $n{=}2$, and $d{=}1024$.
    Gold stars mark the best ACC.
  }
\label{fig:training}
\vspace{-12pt}
\end{figure}

\vspace{-12pt}
\subsubsection{Impact of Aggregation Components}\vspace{-8pt}
Table~\ref{tab:ablation:arch} progressively adds each aggregation component to isolate its individual contributions.
Removing all three components and simply concatenating yields the lowest performance (90.6\% mACC / 96.9\% mAP), confirming that the aggregation design is essential.
BSA alone achieves 95.3\% mACC, demonstrating that directional sequential scanning over the layer hierarchy already provides strong forensic representations.
SWA alone reaches 95.1\% mACC, showing that the softmax-weighted global summary is a competitive alternative but lacks the ordered inductive bias of the SSM.
Combining BSA and SWA improves to 96.1\% mACC, confirming that the two branches capture complementary information, directional sequential evidence, and global layer-wise aggregation, respectively.
Adding SGA to adaptively blend the two outputs further improves to 96.6\% mACC and 99.5\% mAP, validating that the learned scalar gate successfully exploits the complementarity of both branches rather than applying a fixed combination.
\vspace{-12pt}
\subsubsection{Backbone Choice}\vspace{-8pt}
Table~\ref{tab:backbone} compares five backbone architectures under identical training conditions. DINO and CLIP-based backbones achieve competitive but lower performance (91.7--94.8\% mACC), suggesting that their general-purpose visual representations are less well-suited to forensic feature extraction than PE-Core. PE-Core-L14-336 achieves 95.7\% mACC, already outperforming all non-PE alternatives, while scaling to the gigantic PE-Core-G14-448 further improves to 96.6\% mACC and 99.5\% mAP, confirming that both the PE-Core pretraining objective and larger model capacity contribute meaningfully to detection performance.
\vspace{-12pt}
\subsubsection{Projection Configurations and Training Steps}\vspace{-8pt}
\label{sec:steps_proj}
Figure~\ref{fig:training}(a) shows that a single projection layer
($n{=}1$) is insufficient (95.7\% Acc), while two layers ($n{=}2$) achieve
the best result (96.6\% Acc, 99.5\% AP).
Adding further layers ($n{\geq}3$) yields marginal degradation, indicating
that the projection head has enough capacity at $n{=}2$ and deeper stacking
introduces unnecessary parameters.\\
Figure~\ref{fig:training}(b) shows a clear optimum projection dimension at $d{=}1024$
(96.6\% Acc, 99.5\% AP).
Smaller dimensions ($d{=}256$, $d{=}512$) compress the features too
aggressively, losing discriminative information.
Larger dimensions ($d{=}1536$, $d{=}2048$) do not improve accuracy and
increase the parameter count without benefit.
Therefore, we adopt $d{=}1024$ as our default projection dimension for the proposed \methodname{}.\\
As shown in Figure~\ref{fig:training}(c), performance improves
steadily up to 2{,}000 training steps (96.6\% Acc, 99.5\% AP) and degrades slightly
beyond, suggesting the model saturates early, given the low-rank adaptation
regime.
Running longer does not improve generalization and risks overfitting to the
training domain.
We therefore adopt 2{,}000 steps as our default.
\vspace{-12pt}
\subsubsection{Computational Cost}\vspace{-8pt}
\label{sec:ablation:comp_cost}
\begin{wraptable}{r}{0.45\linewidth}
  \vspace{-22pt}
  \centering
  \label{tab:comp_cost}
  \setlength{\tabcolsep}{6pt}
  \begin{tabular}{lc}
    \toprule
    \midrule
    \textbf{Metric}             & \textbf{Value}               \\
    \midrule
    Total parameters            & 1904.32\,M                   \\
    Frozen (backbone)     & 1880.07\,M                   \\
    Trainable (LoRA) & 2.46\,M (0.13\%)             \\
    Trainable (all)       & 24.25\,M (1.3\%)             \\
    GFLOPs                      & 1869.3                       \\
    Inference time              & 62.5\,ms/image  \\
    \midrule
    \bottomrule
  \end{tabular}
  \vspace{6pt}
\caption{Computational cost of the proposed \methodname{} configuration}
\label{tab:comp_cost}
\vspace{-12pt}
\end{wraptable}
Table~\ref{tab:comp_cost} summarizes the computational cost of \methodname{}. The backbone contributes 1{,}880.07\,M parameters, LoRA adds 2.46\,M trainable weights (0.13\%), the full trainable count including the SSM aggregator, projection heads, and gating is 24.25\,M (1.3\%), and the model requires 1{,}869.3 GFLOPs with 62.5\,ms inference time, confirming a highly parameter-efficient and practically deployable design.

\vspace{-12pt}
\section{Conclusions}\vspace{-10pt}
We presented \methodname{}, a novel AIGI detection framework that replaces conventional cross-layer aggregation with three complementary components: a BSA that explicitly models the ordered semantic progression of ViT layer representations through forward and backward selective scans, a SWA that provides a complementary global summary, and an SGA that adaptively fuses both. Built upon a frozen PE-Core backbone with lightweight LoRA adaptation, \methodname{} achieves state-of-the-art performance of 96.6\% mACC and 99.5\% mAP on UniversalFakeDetect and 95.3\% mACC and 98.1\% mAP on AIGCDetect, substantially outperforming the closest predecessor RINE by +5.3\% and +20.5\% mACC, respectively, while maintaining robustness under diverse image perturbations and parameter efficiency with only 1.3\% (0.13\% LoRA only) trainable parameters. These results validate that treating cross-layer ViT features as an ordered sequence and exploiting their directional structure via selective state space modeling is a more powerful and generalizable forensic aggregation strategy than the weighted average.\\
The current framework processes each image independently, without exploiting inter-image relationships, and Midjourney and ADM remain challenging generators whose proprietary and classifier-guided synthesis pipelines introduce artifact profiles not well covered by ProGAN training. Future work will explore multi-source training for broader artifact coverage, extend bidirectional SSM aggregation to video deepfake detection where temporal ordering adds a second sequence dimension, and investigate parallel SSM implementations for improved inference efficiency. \textbf{[Code will be released upon acceptance of the paper.]}

\bibliography{egbib}
\end{document}